\documentclass[letterpaper, 10 pt, conference]{ieeeconf}  

\IEEEoverridecommandlockouts                              

\usepackage{graphicx} 
\usepackage{epsfig} 
\usepackage{mathptmx} 
\usepackage{times} 
\usepackage{amsmath}
\usepackage{bm}
\usepackage{xcolor}
\usepackage{lipsum}
\usepackage{amssymb}  

\usepackage[style=ieee]{biblatex}
\usepackage{booktabs}

\DeclareSourcemap{
  \maps{
    \map{
      \pertype{article}
      \step[fieldset=url, null]
      \step[fieldset=doi, null]
      \step[fieldset=issn, null]
      \step[fieldset=isbn, null]
      \step[fieldset=note, null]
      \step[fieldset=editor, null]
      \step[fieldset=urldate, null]
      \step[fieldset=file, null]
    }
  }
}
\DeclareSourcemap{
  \maps{
    \map{
      \pertype{inproceedings}
      \step[fieldset=url, null]
      \step[fieldset=doi, null]
      \step[fieldset=issn, null]
      \step[fieldset=isbn, null]
      \step[fieldset=note, null]
      \step[fieldset=editor, null]
      \step[fieldset=urldate, null]
      \step[fieldset=file, null]
    }
  }
}
\DeclareSourcemap{
  \maps{
    \map{
      \pertype{incollection}
      \step[fieldset=url, null]
      \step[fieldset=doi, null]
      \step[fieldset=issn, null]
      \step[fieldset=isbn, null]
      \step[fieldset=note, null]
      \step[fieldset=editor, null]
      \step[fieldset=urldate, null]
      \step[fieldset=file, null]
    }
  }
}
\DeclareSourcemap{
  \maps{
    \map{
      \pertype{misc}
      \step[fieldset=url, null]
      \step[fieldset=doi, null]
      \step[fieldset=issn, null]
      \step[fieldset=isbn, null]
      \step[fieldset=note, null]
      \step[fieldset=editor, null]
      \step[fieldset=urldate, null]
      \step[fieldset=file, null]
    }
  }
}
\DeclareSourcemap{
  \maps{
    \map{
      \step[fieldset=url, null]
      \step[fieldset=doi, null]
      \step[fieldset=issn, null]
      \step[fieldset=isbn, null]
      \step[fieldset=note, null]
      \step[fieldset=editor, null]
      \step[fieldset=urldate, null]
      \step[fieldset=file, null]
    }
  }
}
\title{\LARGE \bf
Dynamic Quadrupedal Legged and Aerial Locomotion \\via Structure Repurposing
}

\author{Chenghao Wang$^{1}$, Kaushik Venkatesh Krishnamurthy$^{1}$, Shreyansh Pitroda$^{1}$, Adarsh Salagame$^{1}$, \\Ioannis Mandralis$^{2}$, Eric Sihite$^{2}$, Alireza Ramezani$^{1*}$, and Morteza Gharib$^{2}$
\thanks{$^{1}$ The author is with Department of Electrical and Computer Engineering, Northeastern University, Boston, MA, USA  { wang.chengh, venkateshkrishnamu.k, pitroda.s, a.salagame, a.ramezani@northeastern.edu}}%
\thanks{$^{2}$ The author is with the Department of Aerospace Engineering, California Institute of Technology, Pasadena, CA, USA { esihite, mgharib@caltech.edu}}%
\thanks{$^{*}$ Corresponding author. Email: {a.ramezani@northeastern.edu}}%
}

\begin{document}

\maketitle
\thispagestyle{empty}
\pagestyle{empty}

\begin{abstract}

Multi-modal ground-aerial robots have been extensively studied, with a significant challenge lying in the integration of conflicting requirements across different modes of operation. The Husky robot family, developed at Northeastern University, and specifically the Husky v.2 discussed in this study, addresses this challenge by incorporating posture manipulation and thrust vectoring into multi-modal locomotion through structure repurposing. This quadrupedal robot features leg structures that can be repurposed for dynamic legged locomotion and flight. In this paper, we present the hardware design of the robot and report primary results on dynamic quadrupedal legged locomotion and hovering. 

\end{abstract}

\section{INTRODUCTION}
\label{intro}

Legged robots are intrinsically well-suited for locomotion over difficult terrain \cite{carius_constrained_2022}. These difficulties may be natural environments with rocks, foliage, and soft soil, or city environments with staircases, street curbs, and debris \cite{wermelinger_navigation_2016}. The stepping behavior of the limbs enables locomotion where wheeled locomotion would be impractical, albeit with the drawback of reduced energy efficiency. 
Alternatively, aerial robots \cite{loquercio_learning_2021} 
are well-suited for higher speeds, larger distances, and surmounting obstacles that prohibit all ground locomotion such as waterways, canyons, fences, etc. By proposing a robot design that can morph between legged and aerial mobility, the capabilities of each mode may be harnessed to create a highly versatile robot \cite{sihite_actuation_2023,ramezani_generative_2021}. 

\begin{figure}[h]
    \centering
    \includegraphics[width = \linewidth]{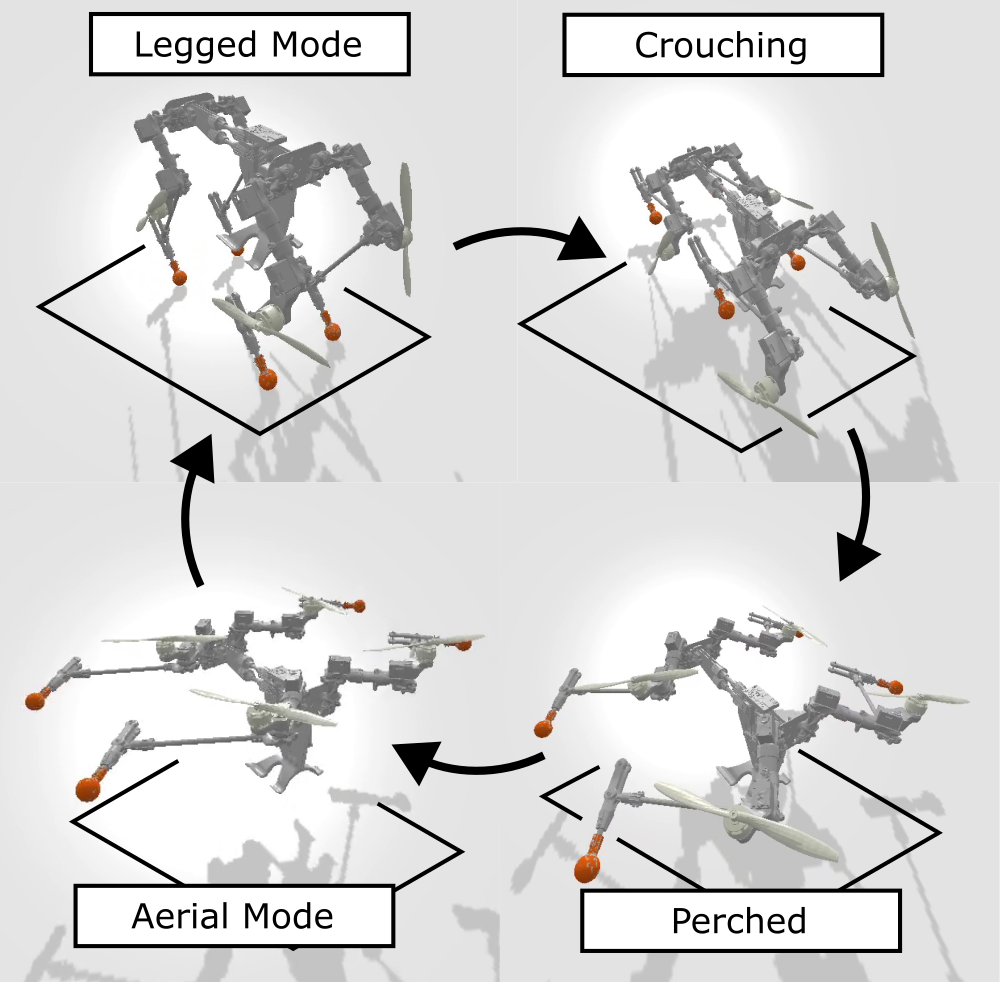}
    \caption{This work explores multi-modal dynamic-legged-aerial locomotion through appendage repurposing.}
    \label{fig:cover-image}
\end{figure}

To showcase the effectiveness of this multi-modal capability, we can look at package delivery problems. These examples are generalized with the purpose of broad applicability. In an automated package delivery mission (in urban or rural environments) \cite{lee_package_2021,stolaroff_energy_2018}, the robot may be deployed from a logistical headquarter several kilometers from the customer. The robot then flies over impassable natural (e.g., rivers, vegetation, hills) or man-made obstacles (e.g., traffic), quickly reaching the vicinity (e.g., building rooftops or balconies) of the customers. There, the robot lands nearby using a powered descent wherever level ground, gaps in the forest canopy, or safe proximity from humans it can find. Now, the legged mobility closes the final distance for the last-mile delivery challenge using its precise and safe legged locomotion capability. 


There are already many promising multi-modal works \cite{fabris_soft_2021,kovac_towards_2010,shin_bio-inspired_2018}. The most common being integrated on the quadrotor platform. But only a few studies exist that show re-configurable systems with appendage and structure repurposing that have demonstrated real-world operability with re-configurable morphologies \cite{luo_design_2014,kim_bipedal_2021, ghassemi_feasibility_2016,pratt_dynamic_2016, shin_development_2019, chukewad_robofly_2021,roderick_bird-inspired_2021}

For instance, LEO \cite{kim_bipedal_2021} can walk and fly. LEO uses thrusters to aid in balancing, which is particularly helpful while climbing over obstacles. The addition of these thrusters also enables agile dynamic tasks such as skateboarding and tightrope walking. There are many variations to LEO such as BALLU \cite{ghassemi_feasibility_2016}, DUCK \cite{pratt_dynamic_2016} that use similar concepts of bipedal locomotion with thrusters. That said, in all of these examples, without thrusters, appendages cannot substantiate terrestrial locomotion which is a disadvantage.
\begin{figure*}[!ht]
    \centering
    \vspace{0.08in}
    \includegraphics[width = \linewidth]{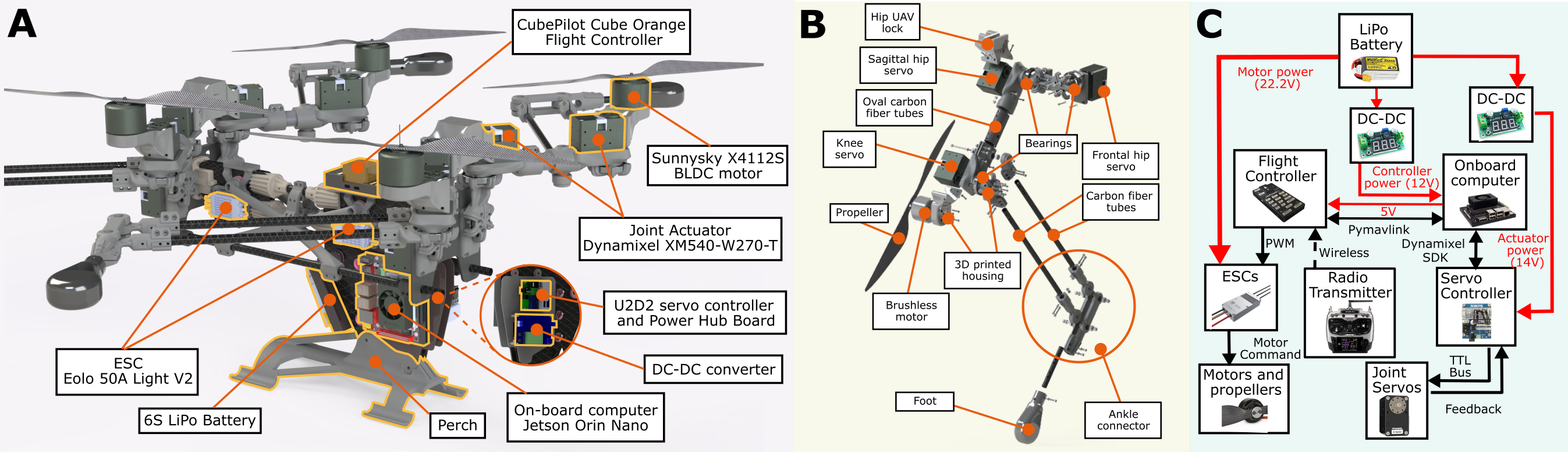}
    \caption{\textbf{A: System Overview}. The system is composed of the main body where the electronics are mounted, and the leg sub-assemblies. The mounting frames and linkages are made out of 3D printed plastic (Markforged Onyx and reinforcing materials) and carbon fiber tubes/plates, respectively.\textbf{ B: Leg Design}. The lower leg, shown as the tibia and fibula linkages, are parallel linkages which is actuated at the knee by the servo. The robot is capable of transforming into the UAV mode and use the propellers mounted at the knee joint for aerial mobility. \textbf{C: Electronics Architecture}: A Lithium Polymer (LiPo) battery powers the entire system, which consists of 4 propeller motors and 12 joint servos. The microcontroller units, which consists of the flight controller and an on-board computing unit, coordinates the input/output of the system to interface with the sensors and controller commands and stabilize the robot.}
    \centering
    \label{fig:husky-beta-overview}
    \vspace{-0.08in}
\end{figure*}
Another common architecture for multi-modal systems seen today is the wheeled rotorcraft. These robots morph between wheeled and aerial mobility. A notable example is Multi-modal Mobility Morphobot (M4) \cite{sihite_multi-modal_2023,sihite_dynamic_2023-1} which has shown extensive locomotion plasticity by showcasing eight distinct modes of operation. 

Many other robots exist that share M4's design principle with active and/or passive wheels and can be used when flying is not possible. For instance, \cite{morton_small_2017,wang_design_2019} and \cite{kalantari_drivocopter_2020} classify and explain different configuration of legged and wheeled aerial systems along with their advantages and downfalls. Their design, called Drivo Copter, features active meshed spherical wheels that surround the propellers and protect the propellers during collision. FCSTAR \cite{david_design_2021} is a re-configurable multi-rotor that showcases multi-modal locomotion. It can fly using propellers and also repurposes the propellers to assist in climbing nearly verticals by providing a reverse thrust. 

The major benefit of these designs is that wheeled mobility has the advantage of fast locomotion and doesn't consume as much power. But where legged motion trumps over wheeled motion is the ability to explore cluttered environments. With the advent of under-actuated passive dynamic walkers, the energy consumption of legged robots can be reduced significantly\cite{kashiri_overview_2018} \cite{ramezani_atrias_2012}. 

Bioinspired robots based on birds and other vertebrates multi-modal operation \cite{dial_wing-assisted_2003-1,tobalske_aerodynamics_2007,peterson_experimental_2011} also exhibit successful multi-modal locomotion. By modeling limbs and actuators similar to animal/human legs, the robot's performance can be significantly improved. Pteromyini \cite{shin_development_2019} is an example of such a multi-modal re-configurable robot, that draws its inspiration from the flying squirrel. The robot uses a membrane between its hind and forelimbs and can sprawl its legs to glide. 

Micro Air and Land Vehicle, MALV, by \cite{bachmann_biologically_2009}, which was intended to be used as a surveillance robot, uses bat wings for gliding and small compliant wheel-legs for moving on the ground. The body of the robot is built using carbon fiber and weighs ~100g, can carry a sensor payload up to 20\% of its weight, can crawl, fly, and land. Deployable Air and Land Explorer Robot (DALER) by \cite{daler_flying_2013} mimics the locomotion of Desmondus Rotundus, also known as the vampire bat. The robot needs to be launched to start flying but exhibits terrestrial and aerial locomotion just using wings. RoboFly\cite{chukewad_robofly_2020} is an insect-sized multi-modal robot that can walk, fly and move over water surfaces. The robot uses a piezoelectric cantilever mechanism to produce a flapping mechanism.

These examples fall short of demonstrating the full capabilities of dynamic legged walkers and aerial systems such as birds \cite{dial_wing-assisted_2003}. The primary contributions of this work are: 

\begin{itemize}
    \item introducing the concept of hardware structure repurposing
    \item full hardware design 
    \item validation and experimentation
\end{itemize}

\noindent for combined dynamic quadrupedal locomotion and hovering flight. The paper is structured as follows: an overview of the robot's mechanical design, ground locomotion tests, a discussion of results, and concluding remarks.

\section{Overview of Husky Carbon v.2 Hardware}
\label{sec:hardware}



We designed Husky v.2 (shown in Fig.~\ref{fig:husky-beta-overview}) with a lighter body structure design and off-the-shelf actuators to fully focus on the controls with a less complex hardware design. Our Husky Carbon v.1 platform \cite{dangol_performance_2020,de_oliveira_thruster-assisted_2020,salagame_quadrupedal_2023,dangol_towards_2020,salagame_letter_2022,sihite_efficient_2022,sihite_multi-modal_2023,ramezani_generative_2021,dangol_control_2021,sihite_unilateral_2021,sihite_optimization-free_2021,pitroda_capture_2024,krishnamurthy_enabling_2024-1,krishnamurthy_narrow-path_2024,krishnamurthy_optimization_2024-1}, which possesses custom design actuators and more comprehensively looks into hardware design aspects, serves as our main platform for multi-modal locomotion and integrated thrust-vectoring and posture manipulations. Husky v.2 mechanical overview and electronics architecture are shown in Figs.~\ref{fig:husky-beta-overview}. In this section, we will discuss the 1) structural design rationale, 2) actuators, and 3) electronics.

\subsection{Design Rationale Based on Structure Repurposing}

\begin{table}[h]
    \centering
    \large 
    \renewcommand{\arraystretch}{1.2} 
    \resizebox{0.95\columnwidth}{!}{ 
    \begin{tabular}{c|c|c|c|c|c|c|c|c|c}
        \toprule
        \rotatebox{30}{\textbf{Body}} & 
        \rotatebox{30}{\textbf{Hip}} & 
        \rotatebox{30}{\textbf{Upper leg}} & 
        \rotatebox{30}{\textbf{Lower leg}} & 
        \rotatebox{30}{\textbf{Ankle}} & 
        \rotatebox{30}{\textbf{Foot}} & 
        \rotatebox{30}{\textbf{BLDC}} & 
        \rotatebox{30}{\textbf{Servos}} & 
        \rotatebox{30}{\textbf{Battery}} & 
        \rotatebox{30}{\textbf{Prop}} \\
        \midrule
        \rotatebox{30}{1.68} & 
        \rotatebox{30}{0.048} & 
        \rotatebox{30}{0.46} & 
        \rotatebox{30}{0.060} & 
        \rotatebox{30}{0.055} & 
        \rotatebox{30}{0.075} & 
        \rotatebox{30}{0.197} & 
        \rotatebox{30}{0.165} & 
        \rotatebox{30}{0.50} & 
        \rotatebox{30}{0.021} \\
        \bottomrule
    \end{tabular}
    }
    \caption{Component Mass Breakdown (kg).}
    \label{tab:part-breakdown}
\end{table}

The design of a robotic system that integrates dynamic legged locomotion with aerial mobility faces a problem known as modal conflicting design requirements. A simple argument, where components are progressively added, can reveal fundamental design challenges, i.e., tradeoffs between adding propulsion for flight and the corresponding increase in mass, which negatively the feasibility of dynamic legged locomotion. Assume mass variations in a design from Steps 1-3 given by 

\begin{equation}
    m_1 = m_b + 2m_L \hspace{1cm} \text{Step-1}
\end{equation}

\noindent where $m_b$ represents the base mass and $m_L$ is the mass of the legs (e.g., 2 legs). As thruster structures are introduced $m_t$, the mass increases to

\begin{equation}
    m_2 = m_1 + 2m_t \hspace{1cm} \text{Step-2}
\end{equation}

\noindent further to 

\begin{equation}
    m_3 = m_2 + 2m_t  \hspace{1cm} \text{Step-3}
\end{equation}

\noindent Observe that in the above design process, the redistribution of leg loading $N$ is involved as more thruster structure are incorporated. Assume the desired leg loading in Step-1 is represented by $N_d$, and its ratio to the actual leg load is given by 

\begin{equation}
    \frac{N_d}{N_1} = \alpha  \hspace{1cm} \text{Step-1}
\end{equation}

\noindent As thruster structures are added, the updated load ratios in each design step can be expressed as 

\begin{equation}
    \frac{N_d}{N_2} = (1 - \frac{2m_t}{m_2}) \alpha  \hspace{1cm} \text{Step-2}
\end{equation}

\begin{equation}
    \frac{N_d}{N_3} = (1 - \frac{4m_t}{m_3}) \alpha  \hspace{1cm} \text{Step-3}
\end{equation}

\noindent which shows that increasing the mass of thruster structures reduces the leg load contribution. This means while adding thrusters aids thrust-to-weight ratio, it can compromise the effectiveness of legged locomotion by reducing leg loading. 

Now, assume the thrust-to-weight ratio for Step-1 $\beta = \frac{T}{mg}$. In the design process, mass increases due to additional thruster structures, the thrust-to-weight ratio must be reconsidered. At an intermediate stage, the modified thrust-to-weight ratio is given by 

\begin{equation}
    \beta' = \left(2 - \frac{4m_t}{m_3} \right) \beta   \hspace{1cm} \text{Step-3}
\end{equation}

\noindent which reveals decrease in $\beta$. The fundamental tradeoff illustrated here revolves around balancing legged versus aerial mobility and can take confounding forms as the design steps increase. 

The issue that was explained above motivates structure repurposing in Husky since by doing it a total mass of 2.8kg (mass of hip, lower-upper leg, ankle and foot summed up and multiplied by 4) will be reused in flight mode, i.e., four-folded thrust-to-weight ratio without added mass. 

The motor-propellers are fixed to each knee joint, allowing the propellers to face upwards when the legs rotate outwards. Figure \ref{fig:husky-beta-overview} shows the components of the leg design. Each leg has three actuated degrees of freedom: hip frontal flexion/extension, hip sagittal flexion/extension, and knee flexion/extension.

The hip sagittal actuator is fixed to the knee actuator via a femur link, which is an off-the-shelf carbon fiber tube with an oval cross-section. The knee actuator drives a four-bar linkage composed of the tibia and fibula links, enabling flexion-extension of the lower leg. To transition to aerial mode, the robot must collapse all four legs and move them sideways simultaneously. 

Ankle connectors, foot linkages, and feet are all 3D-printed using Onyx materials. The use of carbon fiber and fiber-inlay 3D printing with Onyx strengthens the airframe and leg bones without adding excessive weight. This decision sacrifices leg stiffness for reduced weight, which may introduce unwanted compliance, making control more challenging. However, since payload reduction is of considerable importance to us, we accepted the challenge of managing compliant yet lightweight appendages.


\subsection{Electromechanical Components}

To simplify the design for this version, off-the-shelf servomotors are used to actuate each joint in lieu of lighter, more specialized custom hardware that we pursued in \cite{ramezani_generative_2021}. The selected actuators, other than Knee, are Dynamixel XH540-W270-T, which provide a maximum stall torque of 9.9 Nm. For the knee joint, the Dynamixel XM540-W270-T actuator is used, offering up to 10.6 Nm of torque. A combination of the SunnySky X4112S BLDC, EOLO 50A LIGHT ESC, and a 14x4.7 double-blade propeller was employed. This setup, with four such combinations, can generate a maximum thrust of 13.4 kg with a 6s LiPo battery, providing a thruster-to-weight ratio of approximately 2. 

\subsection{Electronics Architecture}

As shown in Fig.~\ref{fig:husky-beta-overview}, the electronics are mounted on two vertical carbon fiber plates in front of the systems to form a front-heavy design. In \cite{ramezani_generative_2021}, we showed through generative design that a front-heavy system can yield a minimized Total Cost of Transport (TCoT) and payload for forward walking. 

The electronics carrier plate is composed of two carbon fiber plates that are spaced from each other at about 1cm using onyx printed spacers. This plate specifically carries electronic speed controllers for thrusters (SunnySky X4112S \& Eolo 50A Light), voltage regulators (Diymore 20A), a Jetson Orin Nano, and Flight controller (Cube Orange+). The battery base is attached on the lower rear side of the main body structure to ensure that the CoM in flight configuration is aligned with the center of lift created by the four propellers. Two lithium-polymer batteries are selected with a nominal voltage of 22.2 V and a mass of 240g. 
\begin{figure*}[t]
    \centering
    \includegraphics[width=\linewidth]{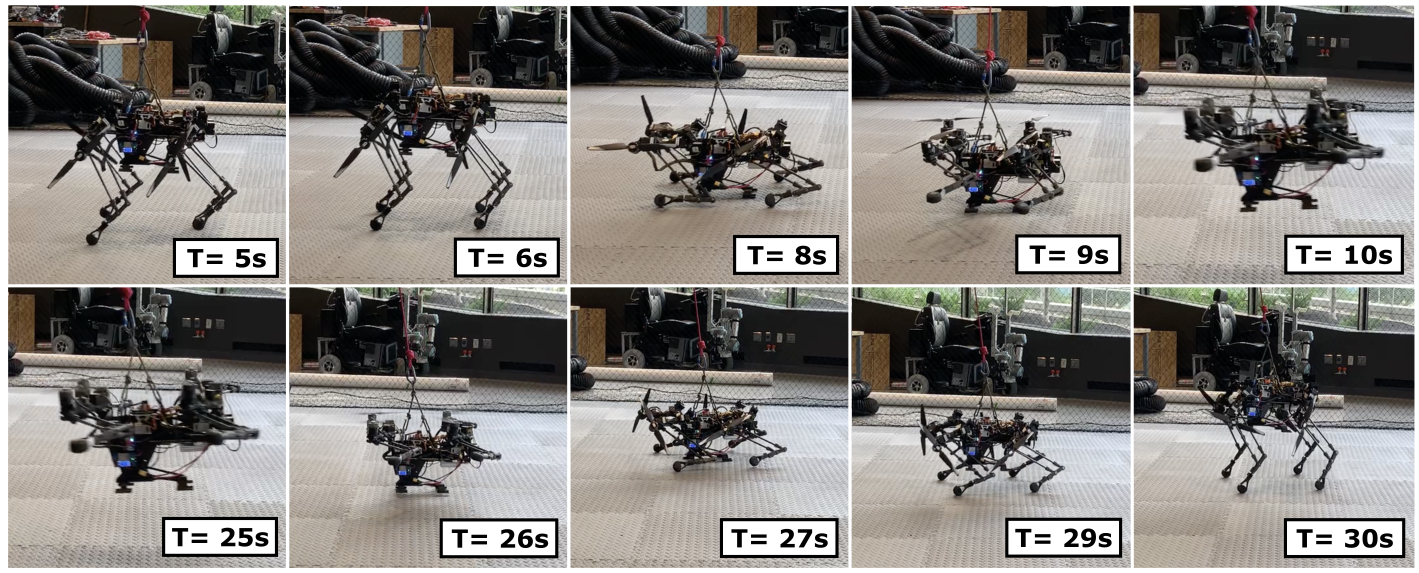}
    \caption{Demonstrates Husky v.2's structure repurposing during transitions between dynamic trotting and hovering, and vice versa.}
    \label{fig:legged-to-aerial}
\end{figure*}
The diagram of the electronic system architecture can be seen in Fig. \ref{fig:husky-beta-overview}. In this diagram, the power cable is depicted as red, showing the division of a 22.2V power source into two distinct outputs through DC-DC converters: 5V for computational power and 12V for actuation power. The flight controller interfaces with the Jetson onboard computer using the pymavlink package and employs PWM signals to communicate with the ESCs. Additionally, the servo controller (U2D2) is operated using the Dynamixel SDK provided by the manufacturer, which manages 12 joint servos through a TTL bus.

\section{Results}
\label{sec:results}

\begin{figure}[t]
    \centering
    \includegraphics[width=0.9\linewidth]{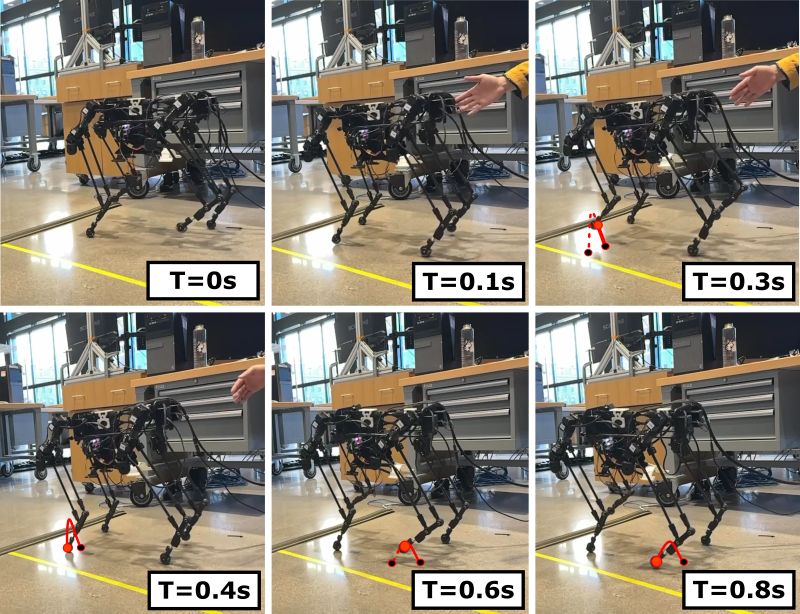}
    \caption{Shows Husky v.2 recovers from a gentle push on its torso during untethered trotting. The leg trajectories are shown as the red lines.}
    \label{fig:trotting-recovery}
\end{figure}
\begin{figure}[h!]
    \centering
    \includegraphics[width = 1.02\linewidth]{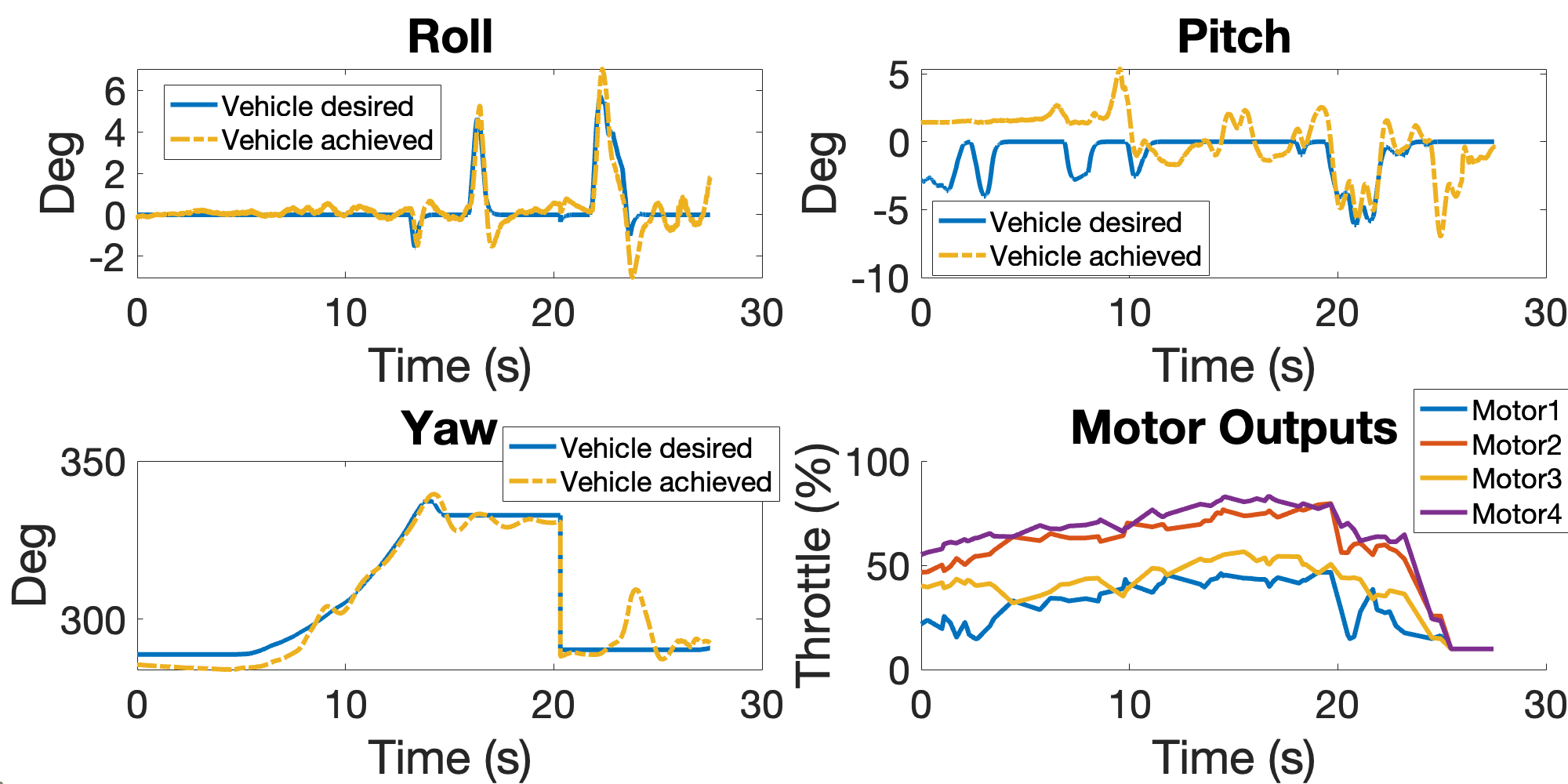}
    \caption{The figure illustrates the vehicle's orientation (roll, pitch, and yaw) angles along with the motor output after the transition to aerial mode. The vehicle initiates takeoff at around the 10-second mark and successfully lands at approximately 25 seconds. The motor output subplot displays the throttle levels, expressed as percentages.}
    \label{fig:rpy-motor}
\end{figure}
\begin{figure}[t]
    \centering  
    \includegraphics[width=\linewidth]{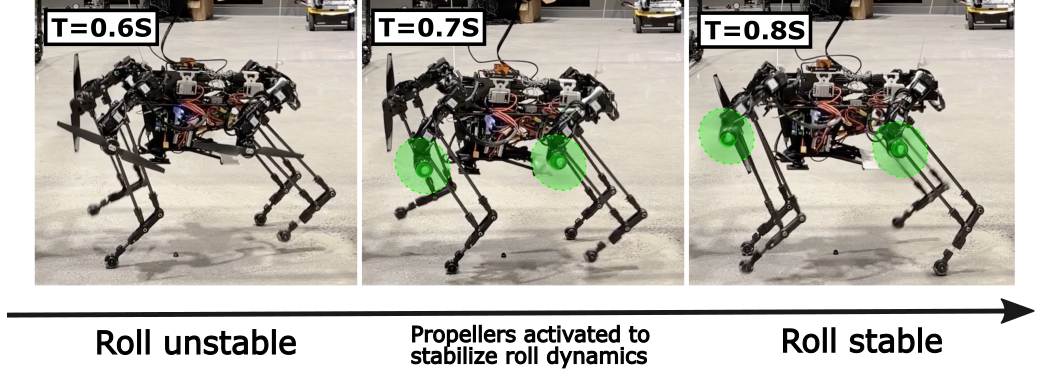}
    \caption{Shows Husky v.2 during untethered trotting while thrusters are active for roll stabilization (dynamic posture manipulation and thrust-vectoring problem).}
    \label{fig:posture-man-thrust-vec}
\end{figure}

\begin{figure}[h!]
    \centering
    \includegraphics[width = 1.02\linewidth]{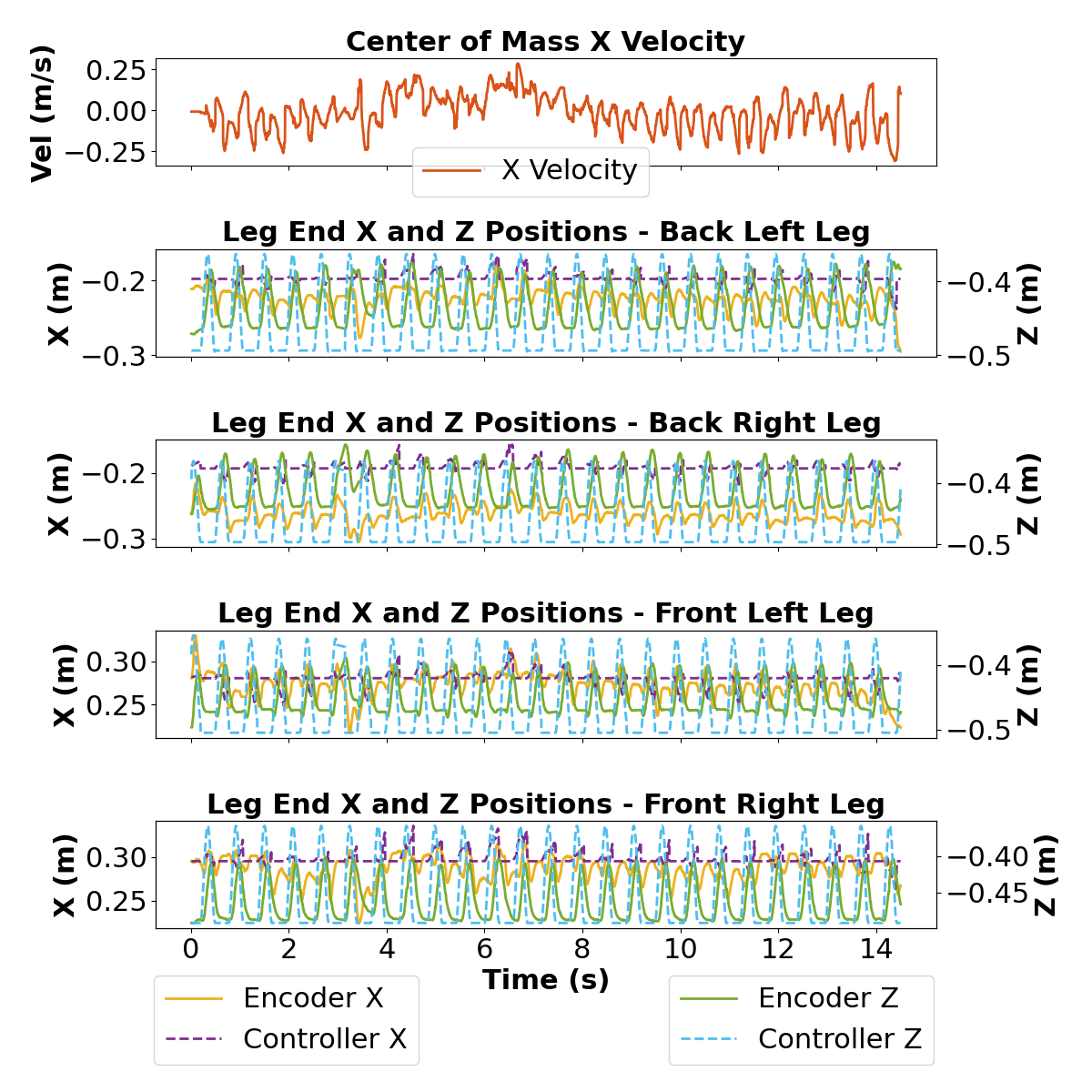}
    \caption{Illustrates the body CoM x-axis velocity, desired, and actual leg ends captured during dynamic trotting.}
    \label{fig:raibert-x}
\end{figure}

\begin{figure}[h!]
    \centering
    \includegraphics[width = 1.05\linewidth]{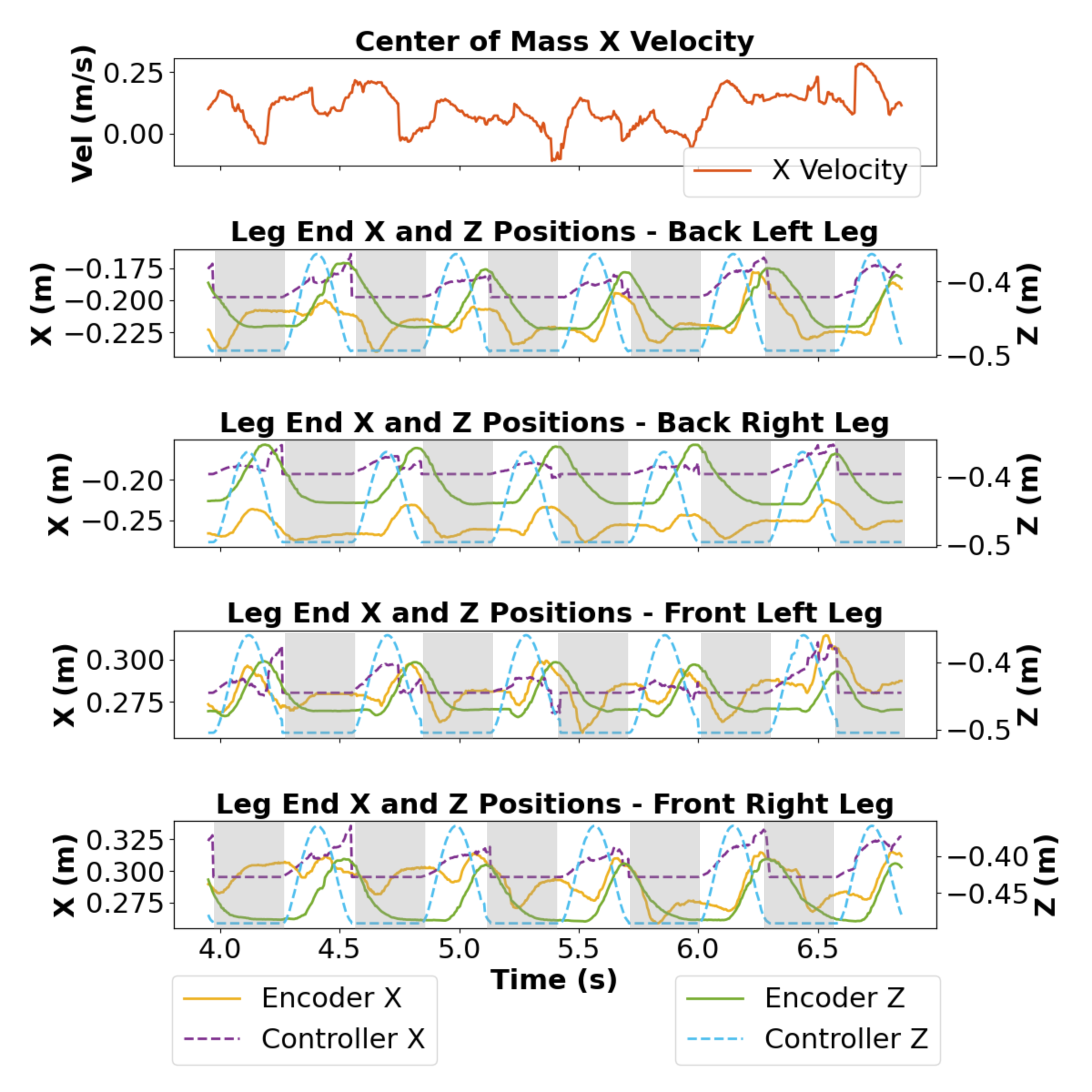}
    \caption{Illustrates the closeup view [4.0-7.0 sec] of body CoM x-axis velocity, desired, and actual leg ends. The stance phase is indicated by the shaded grey regions.}
    \label{fig:raibert-x-closeup}
\end{figure}


We experimentally tested Husky v.2’s untethered dynamic trotting and hovering capabilities. We briefly report these experiments in this section.

In our untethered dynamic trotting experiments, as shown in Figs. \ref{fig:trotting-recovery} (push recovery), \ref{fig:posture-man-thrust-vec}, \ref{fig:raibert-x}, and \ref{fig:raibert-x-closeup} the leading legs lift off the ground and balance about the diagonal axis through the feet of the trailing legs is achieved via foot placement using a Raibert controller. A fourth-order Bézier curve is used to generate the swing trajectory, which has identical first two and last two control points, ensuring zero velocity at both lift-off and touchdown. The joint position commands for individual actuators are then derived using Pinocchio~\cite{carpentier2019pinocchio}, which bases its calculations on the end-effector positions determined by the controller. For hovering flight control, an off-the-shelf controller by Cube Orange+ is used.

The complete pipeline of trotting, transitioning to aerial mode, and hovering is illustrated in Fig.~\ref{fig:legged-to-aerial}. The robot performs dynamic quadrupedal trotting for 8 seconds. The morphing process begins as the robot lowers its body into a crouching position until the perch contacts the ground. The feet continue to lift, transferring the robot’s weight from the feet to the perch. Once the robot is passively balanced on the perch, the transformation from crouching to aerial mode begins.

Next, the legs splay outward via the frontal servos until the femur and tibia are parallel to the floor. Then, the sagittal servos rotate the propellers horizontally until the center of lift aligns with the robot’s center of mass. At this point, the servo positions remain fixed, and the propellers are powered on. The morphing procedure is now considered complete, and the robot adopts a standard quadcopter control scheme. The morphing process takes 10 seconds.

Next, the hovering mode begins, with Husky flying for approximately 20 seconds. The hovering flight is stable in roll, pitch, and yaw, controlled by a human operator. After hovering, the robot lands, followed by the transition back to the legged configuration. Figure.~\ref{fig:rpy-motor} shows flight data.

\section{Concluding Remarks}
\label{sec:conclusion}

In conclusion, we presented the mechanical design and preliminary experimentation of the Husky v.2, a legged-aerial multi-modal morphing robot. Our initial experiments successfully demonstrated the robot's ability to perform:

\begin{itemize}
    \item dynamic quadrupedal trotting (two-contact gait),
    \item gentle push recovery in legged mode using foot placement,
    \item structure repurposing and morphing from legged to aerial modes,
    \item hovering.
\end{itemize}

Our experimental results show that Husky's design possesses thrust-to-weight and leg loading ratios that support both flight and dynamic legged locomotion. While these results suggest that the design concept of structure repurposing is meaningful, other capabilities -- such as carrying extra payloads (e.g., sensors for perception and autonomous navigation) -- remain to be tested and validated.

Hence, our future research will focus on: 1) Demonstrating our robot's ability to leverage its multimodality to negotiate complex environments (e.g., using both legged and flight modes to maneuver around and over obstacles); 2) Developing a unified controller that integrates thruster inputs and ground reaction forces for enhanced thruster-assisted locomotion, including posture manipulation and thrust vectoring for tasks like narrow path walking, slacklining, and steep slope locomotion; 3) Exploring posture manipulation in flight mode by taking advantage of movable propeller arms to recover from propeller fault conditions. 

\section{ACKNOWLEDGMENT}
This work is supported by the Technology Innovation In- stitute (TII), UAE, and the U.S. National Science Foundation (NSF) CAREER Award No. 2340278.

\printbibliography

@inproceedings{daler_flying_2013,
	title = {A Flying Robot with Adaptive Morphology for Multi-Modal Locomotion},
	doi = {10.1109/IROS.2013.6696526},
	eventtitle = {Proceedings of the ... {IEEE}/{RSJ} International Conference on Intelligent Robots and Systems. {IEEE}/{RSJ} International Conference on Intelligent Robots and Systems},
	pages = {1361--1366},
	author = {Daler, Ludovic and Lecoeur, Julien and Hahlen, Patrizia and Floreano, Dario},
	date = {2013-11-01},
}

@article{roderick_bird-inspired_2021,
	title = {Bird-inspired dynamic grasping and perching in arboreal environments},
	rights = {Copyright © 2021 The Authors, some rights reserved; exclusive licensee American Association for the Advancement of Science. No claim to original U.S. Government Works},
	url = {https://www.science.org/doi/abs/10.1126/scirobotics.abj7562},
	doi = {10.1126/scirobotics.abj7562},
	journaltitle = {Science Robotics},
	author = {Roderick, W. R. T. and Cutkosky, M. R. and Lentink, D.},
	urldate = {2022-01-10},
	date = {2021-12-01},
	note = {Publisher: American Association for the Advancement of Science},
}

@article{david_design_2021,
	title = {Design and Analysis of {FCSTAR}, a Hybrid Flying and Climbing Sprawl Tuned Robot},
	volume = {6},
	issn = {2377-3766},
	doi = {10.1109/LRA.2021.3077851},
	pages = {6188--6195},
	number = {4},
	journaltitle = {{IEEE} Robotics and Automation Letters},
	author = {David, Nitzan Ben and Zarrouk, David},
	date = {2021-10},
	note = {Conference Name: {IEEE} Robotics and Automation Letters},
}

@article{wang_design_2019,
	title = {Design and Modeling of a Novel Transformable Land/Air Robot},
	volume = {2019},
	issn = {1687-5966},
	url = {https://www.hindawi.com/journals/ijae/2019/2064131/},
	doi = {10.1155/2019/2064131},
	pages = {e2064131},
	journaltitle = {International Journal of Aerospace Engineering},
	author = {Wang, He and Shi, Jiadong and Wang, Jianzhong and Wang, Hongfeng and Feng, Yiming and You, Yu},
	urldate = {2022-01-10},
	date = {2019-02-04},
	langid = {english},
	note = {Publisher: Hindawi},
}

@inproceedings{kalantari_drivocopter_2020,
	title = {Drivocopter: A concept Hybrid Aerial/Ground vehicle for long-endurance mobility},
	doi = {10.1109/AERO47225.2020.9172782},
	shorttitle = {Drivocopter},
	eventtitle = {2020 {IEEE} Aerospace Conference},
	pages = {1--10},
	booktitle = {2020 {IEEE} Aerospace Conference},
	author = {Kalantari, Arash and Touma, Thomas and Kim, Leon and Jitosho, Rianna and Strickland, Kyle and Lopez, Brett T. and Agha-Mohammadi, Ali-Akbar},
	date = {2020-03},
	note = {{ISSN}: 1095-323X},
}

@inproceedings{morton_small_2017,
	title = {A small hybrid ground-air vehicle concept},
	doi = {10.1109/IROS.2017.8206402},
	eventtitle = {2017 {IEEE}/{RSJ} International Conference on Intelligent Robots and Systems ({IROS})},
	pages = {5149--5154},
	booktitle = {2017 {IEEE}/{RSJ} International Conference on Intelligent Robots and Systems ({IROS})},
	author = {Morton, Scott and Papanikolopoulos, Nikolaos},
	date = {2017-09},
	note = {{ISSN}: 2153-0866},
}

@inproceedings{pratt_dynamic_2016,
	title = {Dynamic underactuated flying-walking ({DUCK}) robot},
	doi = {10.1109/ICRA.2016.7487498},
	eventtitle = {2016 {IEEE} International Conference on Robotics and Automation ({ICRA})},
	pages = {3267--3274},
	booktitle = {2016 {IEEE} International Conference on Robotics and Automation ({ICRA})},
	author = {Pratt, Christopher J. and Leang, Kam K.},
	date = {2016-05},
}

@article{chukewad_robofly_2021,
	title = {{RoboFly}: An Insect-Sized Robot With Simplified Fabrication That Is Capable of Flight, Ground, and Water Surface Locomotion},
	volume = {37},
	issn = {1552-3098, 1941-0468},
	url = {https://ieeexplore.ieee.org/document/9444546/},
	doi = {10.1109/TRO.2021.3075374},
	shorttitle = {{RoboFly}},
	pages = {2025--2040},
	number = {6},
	journaltitle = {{IEEE} Trans. Robot.},
	author = {Chukewad, Yogesh M. and James, Johannes and Singh, Avinash and Fuller, Sawyer},
	urldate = {2022-01-15},
	date = {2021-12},
	langid = {english},
}

@thesis{chukewad_robofly_2020,
	title = {{RoboFly}: Towards Autonomous Flight of a Multimodal Insect-Scale Robot},
	rights = {{CC} {BY}-{NC}},
	url = {https://digital.lib.washington.edu:443/researchworks/handle/1773/46512},
	shorttitle = {{RoboFly}},
	type = {Thesis},
	author = {Chukewad, Yogesh Madhavrao},
	urldate = {2022-01-15},
	date = {2020},
	langid = {american},
	note = {Accepted: 2020-10-26T20:44:05Z},
}

@article{bachmann_biologically_2009,
	title = {A biologically inspired micro-vehicle capable of aerial and terrestrial locomotion},
	volume = {44},
	issn = {0094114X},
	url = {https://linkinghub.elsevier.com/retrieve/pii/S0094114X08001808},
	doi = {10.1016/j.mechmachtheory.2008.08.008},
	pages = {513--526},
	number = {3},
	journaltitle = {Mechanism and Machine Theory},
	author = {Bachmann, Richard J. and Boria, Frank J. and Vaidyanathan, Ravi and Ifju, Peter G. and Quinn, Roger D.},
	urldate = {2022-01-17},
	date = {2009-03},
	langid = {english},
}

@article{luo_design_2014,
	title = {Design and optimization of wheel-legged robot: Rolling-Wolf},
	volume = {27},
	issn = {1000-9345, 2192-8258},
	url = {http://link.springer.com/10.3901/CJME.2014.0905.144},
	doi = {10.3901/CJME.2014.0905.144},
	shorttitle = {Design and optimization of wheel-legged robot},
	pages = {1133--1142},
	number = {6},
	journaltitle = {Chin. J. Mech. Eng.},
	author = {Luo, Yang and Li, Qimin and Liu, Zhangxing},
	urldate = {2022-01-17},
	date = {2014-11},
	langid = {english},
}

@article{shin_development_2019,
	title = {Development and experiments of a bio-inspired robot with multi-mode in aerial and terrestrial locomotion},
	volume = {14},
	issn = {1748-3182, 1748-3190},
	url = {https://iopscience.iop.org/article/10.1088/1748-3190/ab2ab7},
	doi = {10.1088/1748-3190/ab2ab7},
	pages = {056009},
	number = {5},
	journaltitle = {Bioinspir. Biomim.},
	author = {Shin, Won Dong and Park, Jaejun and Park, Hae-Won},
	urldate = {2022-01-19},
	date = {2019-09-01},
	langid = {english},
}

@article{kim_bipedal_2021,
	title = {A bipedal walking robot that can fly, slackline, and skateboard},
	volume = {6},
	url = {https://www.science.org/doi/abs/10.1126/scirobotics.abf8136},
	doi = {10.1126/scirobotics.abf8136},
	pages = {eabf8136},
	number = {59},
	journaltitle = {Science Robotics},
	author = {Kim, Kyunam and Spieler, Patrick and Lupu, Elena-Sorina and Ramezani, Alireza and Chung, Soon-Jo},
	date = {2021},
	note = {\_eprint: https://www.science.org/doi/pdf/10.1126/scirobotics.abf8136},
}

@article{ghassemi_feasibility_2016,
	title = {Feasibility study of a novel robotic system {BALLU}: Buoyancy assisted lightweight legged unit},
	doi = {10.1109/HUMANOIDS.2016.7803268},
	shorttitle = {Feasibility study of a novel robotic system {BALLU}},
	journaltitle = {2016 {IEEE}-{RAS} 16th International Conference on Humanoid Robots (Humanoids)},
	author = {Ghassemi, S. and Hong, D.},
	date = {2016},
}

@article{kashiri_overview_2018,
	title = {An Overview on Principles for Energy Efficient Robot Locomotion},
	volume = {5},
	issn = {2296-9144},
	url = {https://www.frontiersin.org/article/10.3389/frobt.2018.00129},
	journaltitle = {Frontiers in Robotics and {AI}},
	author = {Kashiri, Navvab and Abate, Andy and Abram, Sabrina J. and Albu-Schaffer, Alin and Clary, Patrick J. and Daley, Monica and Faraji, Salman and Furnemont, Raphael and Garabini, Manolo and Geyer, Hartmut and Grabowski, Alena M. and Hurst, Jonathan and Malzahn, Jorn and Mathijssen, Glenn and Remy, David and Roozing, Wesley and Shahbazi, Mohammad and Simha, Surabhi N. and Song, Jae-Bok and Smit-Anseeuw, Nils and Stramigioli, Stefano and Vanderborght, Bram and Yesilevskiy, Yevgeniy and Tsagarakis, Nikos},
	urldate = {2022-02-20},
	date = {2018},
}

@incollection{kovac_towards_2010,
	address = {Berlin, Heidelberg},
	title = {Towards a {Self}-{Deploying} and {Gliding} {Robot}},
	isbn = {978-3-540-89393-6},
	url = {https://doi.org/10.1007/978-3-540-89393-6_19},
	language = {en},
	urldate = {2019-09-04},
	booktitle = {Flying {Insects} and {Robots}},
	publisher = {Springer Berlin Heidelberg},
	author = {Kovač, Mirko and Zufferey, Jean-Christophe and Floreano, Dario},
	editor = {Floreano, Dario and Zufferey, Jean-Christophe and Srinivasan, Mandyam V. and Ellington, Charlie},
	year = {2010},
	doi = {10.1007/978-3-540-89393-6_19},
	pages = {271--284},
}

@incollection{ramezani_atrias_2012,
	title = {Atrias 2.0, a new 3d bipedal robotic walker and runner},
	isbn = {978-981-4415-94-1},
	urldate = {2020-05-09},
	booktitle = {Adaptive {Mobile} {Robotics}},
	publisher = {WORLD SCIENTIFIC},
	author = {Ramezani, Alireza and Grizzle, J.w.},
	month = may,
	year = {2012},
	doi = {10.1142/9789814415958_0060},
	pages = {467--474},
}

@inproceedings{shin_bio-inspired_2018,
	title = {Bio-{Inspired} {Design} of a {Gliding}-{Walking} {Multi}-{Modal} {Robot}},
	doi = {10.1109/IROS.2018.8594210},
	booktitle = {2018 {IEEE}/{RSJ} {International} {Conference} on {Intelligent} {Robots} and {Systems} ({IROS})},
	author = {Shin, Won Dong and Park, Jaejun and Park, Hae-Won},
	month = oct,
	year = {2018},
	pages = {8158--8164},
}

@article{dial_wing-assisted_2003,
	title = {Wing-{Assisted} {Incline} {Running} and the {Evolution} of {Flight}},
	volume = {299},
	url = {https://www.science.org/doi/full/10.1126/science.1078237},
	doi = {10.1126/science.1078237},
	number = {5605},
	urldate = {2022-10-03},
	journal = {Science},
	author = {Dial, Kenneth P.},
	month = jan,
	year = {2003},
	pages = {402--404},
}

@article{tobalske_aerodynamics_2007,
	title = {Aerodynamics of wing-assisted incline running in birds},
	volume = {210},
	issn = {0022-0949},
	url = {https://doi.org/10.1242/jeb.001701},
	doi = {10.1242/jeb.001701},
	number = {10},
	urldate = {2022-10-03},
	journal = {Journal of Experimental Biology},
	author = {Tobalske, Bret W. and Dial, Kenneth P.},
	month = may,
	year = {2007},
	pages = {1742--1751},
}

@article{dial_wing-assisted_2003-1,
	title = {Wing-{Assisted} {Incline} {Running} and the {Evolution} of {Flight}},
	volume = {299},
	url = {https://www.science.org/doi/10.1126/science.1078237},
	doi = {10.1126/science.1078237},
	number = {5605},
	urldate = {2022-10-05},
	journal = {Science},
	author = {Dial, Kenneth P.},
	month = jan,
	year = {2003},
	pages = {402--404},
}

@article{stolaroff_energy_2018,
	title = {Energy use and life cycle greenhouse gas emissions of drones for commercial package delivery},
	volume = {9},
	copyright = {2018 The Author(s)},
	issn = {2041-1723},
	url = {https://www.nature.com/articles/s41467-017-02411-5},
	doi = {10.1038/s41467-017-02411-5},
	language = {en},
	number = {1},
	urldate = {2022-10-08},
	journal = {Nature Communications},
	author = {Stolaroff, Joshuah K. and Samaras, Constantine and O’Neill, Emma R. and Lubers, Alia and Mitchell, Alexandra S. and Ceperley, Daniel},
	month = feb,
	year = {2018},
	pages = {409},
}

@inproceedings{lee_package_2021,
	address = {New York, NY, USA},
	series = {{UbiComp} '21},
	title = {Package {Delivery} {Using} {Autonomous} {Drones} in {Skyways}},
	isbn = {978-1-4503-8461-2},
	url = {https://doi.org/10.1145/3460418.3479289},
	doi = {10.1145/3460418.3479289},
	urldate = {2022-10-23},
	booktitle = {Adjunct {Proceedings} of the 2021 {ACM} {International} {Joint} {Conference} on {Pervasive} and {Ubiquitous} {Computing} and {Proceedings} of the 2021 {ACM} {International} {Symposium} on {Wearable} {Computers}},
	publisher = {Association for Computing Machinery},
	author = {Lee, Woojin and Alkouz, Balsam and Shahzaad, Babar and Bouguettaya, Athman},
	month = sep,
	year = {2021},
	pages = {48--50},
}

@inproceedings{fabris_soft_2021,
	title = {A {Soft} {Drone} with {Multi}-modal {Mobility} for the {Exploration} of {Confined} {Spaces}},
	doi = {10.1109/SSRR53300.2021.9597683},
	booktitle = {2021 {IEEE} {International} {Symposium} on {Safety}, {Security}, and {Rescue} {Robotics} ({SSRR})},
	author = {Fabris, Amedeo and Kirchgeorg, Steffen and Mintchev, Stefano},
	month = oct,
	year = {2021},
	pages = {48--54},
}

@inproceedings{peterson_experimental_2011,
	title = {Experimental dynamics of wing assisted running for a bipedal ornithopter},
	doi = {10.1109/IROS.2011.6095041},
	booktitle = {2011 {IEEE}/{RSJ} {International} {Conference} on {Intelligent} {Robots} and {Systems}},
	author = {Peterson, Kevin and Fearing, Ronald S.},
	month = sep,
	year = {2011},
	pages = {5080--5086},
}

@article{loquercio_learning_2021,
	title = {Learning high-speed flight in the wild},
	volume = {6},
	url = {https://www.science.org/doi/full/10.1126/scirobotics.abg5810},
	doi = {10.1126/scirobotics.abg5810},
	number = {59},
	urldate = {2023-09-26},
	journal = {Science Robotics},
	author = {Loquercio, Antonio and Kaufmann, Elia and Ranftl, René and Müller, Matthias and Koltun, Vladlen and Scaramuzza, Davide},
	month = oct,
	year = {2021},
	pages = {eabg5810},
}

@misc{salagame_letter_2022,
	title = {A {Letter} on {Progress} {Made} on {Husky} {Carbon}: {A} {Legged}-{Aerial}, {Multi}-modal {Platform}},
	shorttitle = {A {Letter} on {Progress} {Made} on {Husky} {Carbon}},
	url = {http://arxiv.org/abs/2207.12254},
	urldate = {2023-05-17},
	publisher = {arXiv},
	author = {Salagame, Adarsh and Manjikian, Shoghair and Wang, Chenghao and Krishnamurthy, Kaushik Venkatesh and Pitroda, Shreyansh and Gupta, Bibek and Jacob, Tobias and Mottis, Benjamin and Sihite, Eric and Ramezani, Milad and Ramezani, Alireza},
	month = jul,
	year = {2022},
	doi = {10.48550/arXiv.2207.12254},
}

@inproceedings{sihite_actuation_2023,
	title = {Actuation and {Flight} {Control} of {High}-{DOF} {Dynamic} {Morphing} {Wing} {Flight} by {Shifting} {Structure} {Response}},
	url = {https://ieeexplore.ieee.org/document/10383886},
	doi = {10.1109/CDC49753.2023.10383886},
	urldate = {2024-08-26},
	booktitle = {2023 62nd {IEEE} {Conference} on {Decision} and {Control} ({CDC})},
	author = {Sihite, Eric and Salagame, Adarsh and Ghanem, Paul and Ramezani, Alireza},
	month = dec,
	year = {2023},
	pages = {8824--8829},
}

@inproceedings{pitroda_capture_2024,
	title = {Capture {Point} {Control} in {Thruster}-{Assisted} {Bipedal} {Locomotion}},
	url = {https://ieeexplore.ieee.org/document/10637139},
	doi = {10.1109/AIM55361.2024.10637139},
	urldate = {2024-08-26},
	booktitle = {2024 {IEEE} {International} {Conference} on {Advanced} {Intelligent} {Mechatronics} ({AIM})},
	author = {Pitroda, Shreyansh and Bondada, Aditya and Venkatesh, Kaushik and Salagame, Adarsh and Wang, Chenghao and Liu, Taoran and Gupta, Bibek and Sihite, Eric and Nemovi, Reza and Ramezani, Alireza and Gharib, Morteza},
	month = jul,
	year = {2024},
	pages = {1139--1144},
}

@article{dangol_control_2021,
	title = {Control of {Thruster}-{Assisted}, {Bipedal} {Legged} {Locomotion} of the {Harpy} {Robot}},
	volume = {8},
	issn = {2296-9144},
	url = {https://www.frontiersin.org/articles/10.3389/frobt.2021.770514},
	urldate = {2023-05-17},
	journal = {Frontiers in Robotics and AI},
	author = {Dangol, Pravin and Sihite, Eric and Ramezani, Alireza},
	year = {2021},
}

@inproceedings{sihite_efficient_2022,
	title = {Efficient {Path} {Planning} and {Tracking} for {Multi}-{Modal} {Legged}-{Aerial} {Locomotion} {Using} {Integrated} {Probabilistic} {Road} {Maps} ({PRM}) and {Reference} {Governors} ({RG})},
	doi = {10.1109/CDC51059.2022.9992754},
	booktitle = {2022 {IEEE} 61st {Conference} on {Decision} and {Control} ({CDC})},
	author = {Sihite, Eric and Mottis, Benjamin and Ghanem, Paul and Ramezani, Alireza and Gharib, Morteza},
	month = dec,
	year = {2022},
	pages = {764--770},
}

@inproceedings{ramezani_generative_2021,
	title = {Generative {Design} of {NU}’s {Husky} {Carbon}, {A} {Morpho}-{Functional}, {Legged} {Robot}},
	url = {https://ieeexplore.ieee.org/abstract/document/9561196},
	doi = {10.1109/ICRA48506.2021.9561196},
	urldate = {2023-11-22},
	booktitle = {2021 {IEEE} {International} {Conference} on {Robotics} and {Automation} ({ICRA})},
	author = {Ramezani, Alireza and Dangol, Pravin and Sihite, Eric and Lessieur, Andrew and Kelly, Peter},
	month = may,
	year = {2021},
	pages = {4040--4046},
}

@inproceedings{sihite_optimization-free_2021,
	title = {Optimization-free {Ground} {Contact} {Force} {Constraint} {Satisfaction} in {Quadrupedal} {Locomotion}},
	doi = {10.1109/CDC45484.2021.9683155},
	booktitle = {2021 60th {IEEE} {Conference} on {Decision} and {Control} ({CDC})},
	author = {Sihite, Eric and Dangol, Pravin and Ramezani, Alireza},
	month = dec,
	year = {2021},
	pages = {713--719},
}

@inproceedings{krishnamurthy_narrow-path_2024,
	title = {Narrow-{Path}, {Dynamic} {Walking} {Using} {Integrated} {Posture} {Manipulation} and {Thrust} {Vectoring}},
	url = {https://ieeexplore.ieee.org/document/10637015},
	doi = {10.1109/AIM55361.2024.10637015},
	urldate = {2024-08-26},
	booktitle = {2024 {IEEE} {International} {Conference} on {Advanced} {Intelligent} {Mechatronics} ({AIM})},
	author = {Krishnamurthy, Kaushik Venkatesh and Wang, Chenghao and Pitroda, Shreyansh and Salagame, Adarsh and Sihite, Eric and Nemovi, Reza and Ramezani, Alireza and Gharib, Morteza},
	month = jul,
	year = {2024},
	pages = {898--903},
}

@inproceedings{dangol_performance_2020,
	title = {Performance satisfaction in {Midget}, a thruster-assisted bipedal robot},
	doi = {10.23919/ACC45564.2020.9147448},
	booktitle = {2020 {American} {Control} {Conference} ({ACC})},
	author = {Dangol, Pravin and Ramezani, Alireza and Jalili, Nader},
	month = jul,
	year = {2020},
	pages = {3217--3223},
}

@article{sihite_multi-modal_2023,
	title = {Multi-{Modal} {Mobility} {Morphobot} ({M4}) with appendage repurposing for locomotion plasticity enhancement},
	volume = {14},
	copyright = {2023 Springer Nature Limited},
	issn = {2041-1723},
	url = {https://www.nature.com/articles/s41467-023-39018-y},
	doi = {10.1038/s41467-023-39018-y},
	language = {en},
	number = {1},
	urldate = {2023-10-07},
	journal = {Nature Communications},
	author = {Sihite, Eric and Kalantari, Arash and Nemovi, Reza and Ramezani, Alireza and Gharib, Morteza},
	month = jun,
	year = {2023},
	pages = {3323},
}

@inproceedings{sihite_unilateral_2021,
	title = {Unilateral {Ground} {Contact} {Force} {Regulations} in {Thruster}-{Assisted} {Legged} {Locomotion}},
	url = {https://ieeexplore.ieee.org/abstract/document/9517648},
	doi = {10.1109/AIM46487.2021.9517648},
	urldate = {2023-11-22},
	booktitle = {2021 {IEEE}/{ASME} {International} {Conference} on {Advanced} {Intelligent} {Mechatronics} ({AIM})},
	author = {Sihite, Eric and Dangol, Pravin and Ramezani, Alireza},
	month = jul,
	year = {2021},
	pages = {389--395},
}

@inproceedings{de_oliveira_thruster-assisted_2020,
	title = {Thruster-assisted {Center} {Manifold} {Shaping} in {Bipedal} {Legged} {Locomotion}},
	doi = {10.1109/AIM43001.2020.9158967},
	booktitle = {2020 {IEEE}/{ASME} {International} {Conference} on {Advanced} {Intelligent} {Mechatronics} ({AIM})},
	author = {de Oliveira, Arthur C. B. and Ramezani, Alireza},
	month = jul,
	year = {2020},
	pages = {508--513},
}

@article{dangol_towards_2020,
	series = {21st {IFAC} {World} {Congress}},
	title = {Towards thruster-assisted bipedal locomotion for enhanced efficiency and robustness},
	volume = {53},
	issn = {2405-8963},
	url = {https://www.sciencedirect.com/science/article/pii/S2405896320334844},
	doi = {10.1016/j.ifacol.2020.12.2721},
	language = {en},
	number = {2},
	urldate = {2023-05-17},
	journal = {IFAC-PapersOnLine},
	author = {Dangol, Pravin and Ramezani, Alireza},
	month = jan,
	year = {2020},
	pages = {10019--10024},
}

@misc{salagame_quadrupedal_2023,
	title = {Quadrupedal {Locomotion} {Control} {On} {Inclined} {Surfaces} {Using} {Collocation} {Method}},
	url = {http://arxiv.org/abs/2312.08621},
	urldate = {2024-01-17},
	publisher = {arXiv},
	author = {Salagame, Adarsh and Gianello, Maria and Wang, Chenghao and Venkatesh, Kaushik and Pitroda, Shreyansh and Rajput, Rohit and Sihite, Eric and Leeser, Miriam and Ramezani, Alireza},
	month = dec,
	year = {2023},
	doi = {10.48550/arXiv.2312.08621},
}

@misc{krishnamurthy_optimization_2024-1,
	title = {Optimization free control and ground force estimation with momentum observer for a multimodal legged aerial robot},
	url = {http://arxiv.org/abs/2411.11216},
	doi = {10.48550/arXiv.2411.11216},
	urldate = {2025-03-12},
	publisher = {arXiv},
	author = {Krishnamurthy, Kaushik Venkatesh and Wang, Chenghao and Pitroda, Shreyansh and Sihite, Eric and Ramezani, Alireza and Gharib, Morteza},
	month = nov,
	year = {2024},
	note = {arXiv:2411.11216 [cs]},
}

@misc{krishnamurthy_enabling_2024-1,
	title = {Enabling steep slope walking on {Husky} using reduced order modeling and quadratic programming},
	url = {http://arxiv.org/abs/2411.11788},
	doi = {10.48550/arXiv.2411.11788},
	urldate = {2025-03-12},
	publisher = {arXiv},
	author = {Krishnamurthy, Kaushik Venkatesh and Sihite, Eric and Wang, Chenghao and Pitroda, Shreyansh and Salagame, Adarsh and Ramezani, Alireza and Gharib, Morteza},
	month = nov,
	year = {2024},
	note = {arXiv:2411.11788 [cs]},
}

@article{carius_constrained_2022,
	title = {Constrained stochastic optimal control with learned importance sampling: {A} path integral approach},
	volume = {41},
	issn = {0278-3649},
	shorttitle = {Constrained stochastic optimal control with learned importance sampling},
	url = {https://doi.org/10.1177/02783649211047890},
	doi = {10.1177/02783649211047890},
	language = {en},
	number = {2},
	urldate = {2025-03-01},
	journal = {The International Journal of Robotics Research},
	author = {Carius, Jan and Ranftl, René and Farshidian, Farbod and Hutter, Marco},
	month = feb,
	year = {2022},
	note = {Publisher: SAGE Publications Ltd STM},
	pages = {189--209},
}

@inproceedings{wermelinger_navigation_2016,
	title = {Navigation planning for legged robots in challenging terrain},
	url = {https://ieeexplore.ieee.org/document/7759199/?arnumber=7759199},
	doi = {10.1109/IROS.2016.7759199},
	urldate = {2025-03-01},
	booktitle = {2016 {IEEE}/{RSJ} {International} {Conference} on {Intelligent} {Robots} and {Systems} ({IROS})},
	author = {Wermelinger, Martin and Fankhauser, Péter and Diethelm, Remo and Krüsi, Philipp and Siegwart, Roland and Hutter, Marco},
	month = oct,
	year = {2016},
	note = {ISSN: 2153-0866},
	pages = {1184--1189},
}

@misc{sihite_dynamic_2023-1,
	title = {Dynamic modeling of wing-assisted inclined running with a morphing multi-modal robot},
	url = {http://arxiv.org/abs/2311.09963},
	doi = {10.48550/arXiv.2311.09963},
	urldate = {2023-12-02},
	publisher = {arXiv},
	author = {Sihite, Eric and Ramezani, Alireza and Gharib, Morteza},
	month = nov,
	year = {2023},
	note = {arXiv:2311.09963 [cs, eess]},
}

@inproceedings{carpentier2019pinocchio,
   title={The Pinocchio C++ library -- A fast and flexible implementation of rigid body dynamics algorithms and their analytical derivatives},
   author={Carpentier, Justin and Saurel, Guilhem and Buondonno, Gabriele and Mirabel, Joseph and Lamiraux, Florent and Stasse, Olivier and Mansard, Nicolas},
   booktitle={IEEE International Symposium on System Integrations (SII)},
   year={2019}
}

\end{document}